\documentclass[journal=jacsat,manuscript=article]{achemso}

\usepackage[version=3]{mhchem} % Formula subscripts using \ce{}
\usepackage{graphicx}
\usepackage{threeparttable}
\usepackage{hyperref}

\author{Peter Pak}
\affiliation{
  Department of Mechanical Engineering, Carnegie Mellon University, Pittsburgh,
  PA, USA
}
\author{Amir Barati Farimani}
\email{barati@cmu.edu}
\affiliation{
  Department of Mechanical Engineering, Carnegie Mellon University, Pittsburgh,
  PA, USA
}
\alsoaffiliation{
  Machine Learning Department, Carnegie Mellon University, Pittsburgh, PA, USA
}

\title[]{
    AdditiveLLM: Large Language Models Predict Defects in Metals Additive
    Manufacturing
}

\abbreviations{IR,NMR,UV}

\SectionNumbersOn
\begin{document}

\begin{abstract}
In this work we investigate the ability of large language models to predict
additive manufacturing defect regimes given a set of process parameter inputs.
For this task we utilize a process parameter defect dataset to fine-tune a
collection of models, titled \textit{AdditiveLLM}, for the purpose of predicting
potential defect regimes including \textit{Keyholing}, \textit{Lack of Fusion},
and \textit{Balling}. We compare different methods of input formatting in order
to gauge the model's performance to correctly predict defect regimes on our
sparse \textit{Baseline} dataset and our natural language \textit{Prompt}
dataset. The model displays robust predictive capability, achieving an accuracy
of 93\% when asked to provide the defect regimes associated with a set of
process parameters. The incorporation of natural language input further
simplifies the task of process parameters selection, enabling users to identify
optimal settings specific to their build.

\end{abstract}

\section{Introduction}
Within the process of Laser Powder Bed Fusion (L-PBF), the selection of optimal
process parameters remains a key factor in the fabrication of defect free parts
\cite{beuth_process_2013, gordon_defect_2020}. This is achieved by avoiding the
combination of process parameters that can potentially lead to melt pool
characteristics resulting in either unfused powder or void initiation causing
defect formation within the final part. The following defect regimes of
\textit{Keyholing}, \textit{Balling}, and \textit{Lack of Fusion (LoF)} describe the
expected melt pool characteristics seen at a given combination of process
parameters \cite{gordon_defect_2020}. These defect regimes can be estimated by
methods such as experimental observation \cite{ahmed_process_2022},
computational simulation \cite{roh_ontology-based_2021}, or surrogate modeling
\cite{menon_multi-fidelity_2022}.

Equations \ref{eq:keyhole} - \ref{eq:balling} provide criterions dependent on
melt pool dimensions and process parameters where when satisfied are expected to
avoid their respective defect regimes. Melt pool conditions such as
\textit{Keyholing} do not inherently cause defects, rather it is fluctuations
and collapse of the rear wall that generates voids leading to the formation of
pores \cite{huang_keyhole_2022}. However due to the inaccessibility of
\textit{in-situ} cross-sectional melt pool analysis, the formation of
\textit{Keyhole} defects is often approximated with Equation \ref{eq:keyhole},
where width to depth ratios below threshold are likely to result in keyhole
porosity\cite{zhang_efficient_2021}. For defects resulting from \textit{Lack of
Fusion}, the ratio of hatch spacing and melt pool width along with layer height
and depth are evaluated to ascertain the potential formation of porosity
\cite{tang_prediction_2017}. The criterion for full melting across subsequent
melt pools is defined as Equation \ref{eq:lof} where computed values above the
threshold can result in unfused powder \cite{tang_prediction_2017,
miner_impact_2024}. \textit{Balling} defects occur due to the hydrodynamic
capillary instabilities that occur during high scan speeds and the resulting
grooves along the sides of the melt pool result in formation of voids if not
remelted \cite{gu_balling_2007}. An approximation for the occurrence of this
phenomenon is calculated by comparing the length and width of the melt pool
(Eq. \ref{eq:balling}). A ratio above $\pi$ can indicate the presence of
balling, though this value can range depending on the material
\cite{zhang_efficient_2021}.

% Width to depth ratio must be greater than 1.5 to NOT be in keyhole
\begin{equation}
\frac{Width}{Depth} > 1.5
\label{eq:keyhole}
\end{equation}

% Hatch Spacing and Layer Height ratios must be less that 1 to NOT have lack of fusion defects
\begin{equation}
\left(\frac{Hatch\;Spacing}{Width}\right)^2 + \left(\frac{Layer\;Height}{Depth}\right)^2 \le 1
\label{eq:lof}
\end{equation}

% Length to width ratio must be less than \pi to NOT be balling
\begin{equation}
\frac{Length}{Width} < \pi
\label{eq:balling}
\end{equation}

The selection of optimal process parameters often requires extensive domain
knowledge of the various domains within the L-PBF process, ranging from material
properties to experimental trials. Large Language Models (LLM) present a
suitable solution for encapsulating these domains and have demonstrated
significant success in reasoning and generating precise responses based on a
given prompt \cite{touvron_llama_2023, touvron_llama_2023-1, minaee_large_2024,
devlin_bert_2019, brown_language_2020, liu_roberta_2019}. LLMs are able to
understand textual descriptions of experimental results and respective context
which enables its application as a more human accessible research tool in a wide
domain of fields ranging from chemical engineering to robotics
\cite{jadhav_llm-3d_2024, ock_catalyst_2023, car_plato_2024,
lorsung_explain_2024, shah_peptide-gpt_2024, farimani_large_2024,
chandrasekhar_amgpt_2024, ock_multimodal_2024, balaji_gpt-molberta_2023,
jadhav_large_2024, ock_adsorb-agent_2024}. Implementations such as
LLAMA\cite{touvron_llama_2023, touvron_llama_2023-1, grattafiori_llama_2024},
GPT\cite{brown_language_2020}, T5\cite{raffel_exploring_2023} and
BERT\cite{devlin_bert_2019} provide a foundation for general purpose use where
it can then be fine-tuned for a specific downstream application.

The Bidirectional Encoder Representations from Transformers
(BERT)\cite{devlin_bert_2019} model is one way to architect a LLM and has
garnered much attention creating the base for extensions such as RoBERTa
(\textbf{R}obustly \textbf{o}ptimized \textbf{BERT}
Approach)\cite{liu_roberta_2019}, ALBERT (\textbf{A} \textbf{L}ite
\textbf{BERT})\cite{lan_albert_2020}, and DistilBERT\cite{sanh_distilbert_2020}.
These models which improves upon the existing implementation by either employing
specific training strategies such as a dataset, batch sizes, sequence lengths,
and hyperparameter optimization\cite{liu_roberta_2019, lan_albert_2020,
sanh_distilbert_2020}. The BERT model differs from the classical masked language
model in that it randomly masks its input tokens in order to predict the
original vocabulary based solely through context\cite{devlin_bert_2019}. The
bidirectional approach for representing inputs that BERT takes allows for the
model to consider a more complete context window of a given sequence, providing
tokens on both sides of an applied mask \cite{devlin_bert_2019}. This is
contrast to the unidirectional approach that models such as GPT take in order to
more align closer to the sequential nature of text generation
\cite{radford_improving_nodate}.

In this work we propose AdditiveLLM, a comparison of large language models
fine-tuned on experimental and simulation based L-PBF melt pool data that able
to predict potential defects given a combination of process parameters. We focus
upon fine-tuning several models including:
DistilBERT\cite{sanh_distilbert_2020}, SciBERT\cite{beltagy_scibert_2019}, LLama
3.2\cite{noauthor_meta-llamallama-32-1b_2024}, and
T5\cite{noauthor_google-t5t5-small_2024}. Our selection of models utilize the
smaller parameter set variants as this best suits our need for an efficient and
lightweight model for specific inferences regarding defect regimes in L-PBF
Additive Manufacturing. This approach infers upon the corpus of L-PBF melt pool
data formatted as a set of process parameters inputs and defect classification
labels. The model predictions provide a classification to a combination of
potential defect regimes of \textit{Keyholing}, \textit{Lack of Fusion},
\textit{Balling}, and \textit{None} (for the case where no defects exist).
Through inference the models are able to quickly provide insight on potential
defects given a specific parameter combination without the need to perform
additional simulations or experimental builds.

\begin{figure}[ht]
  \centering
  \includegraphics[width=\textwidth]{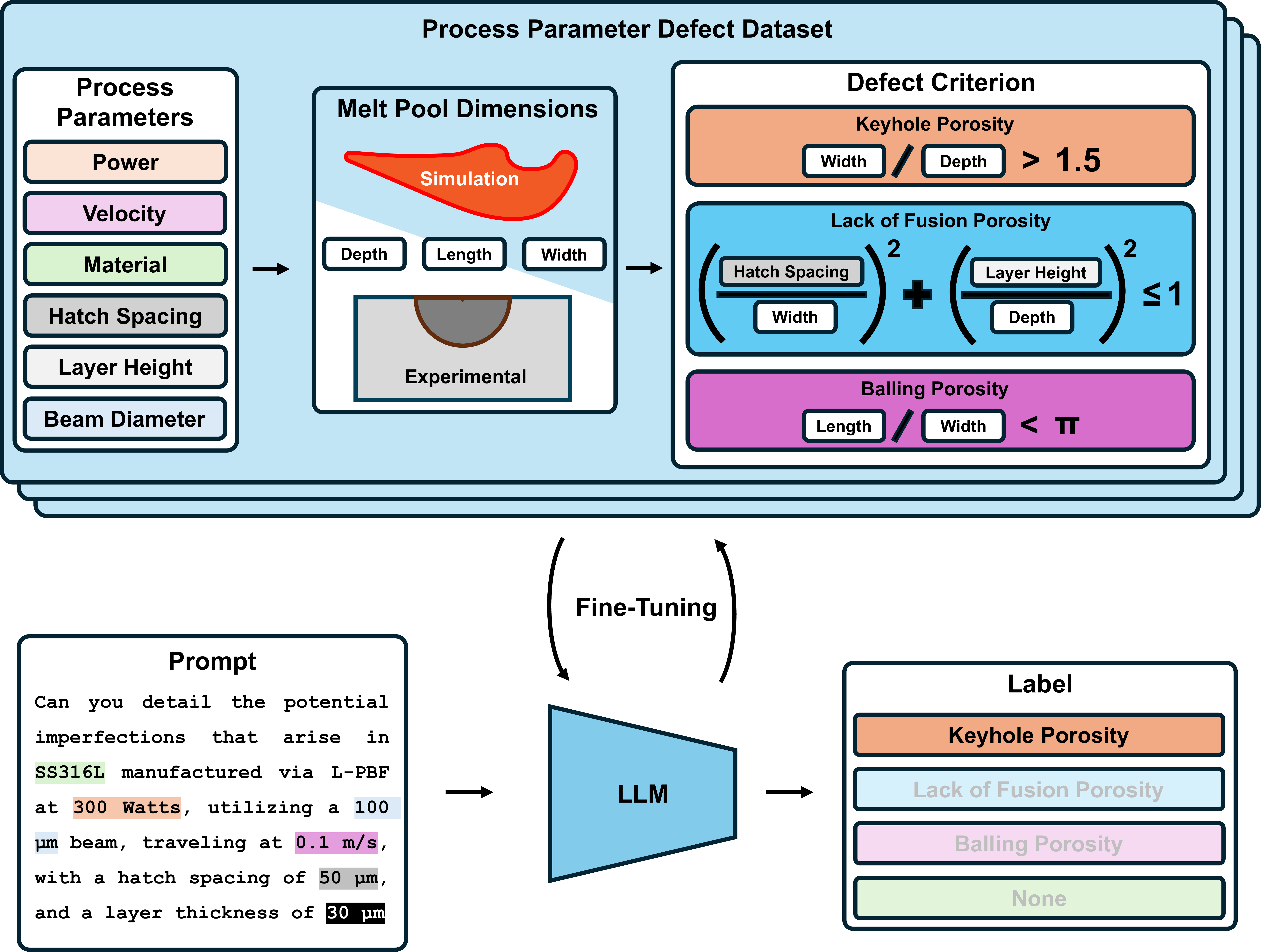}
  \caption{
    A series of LLM models are fine-tuned with a process parameter defect
    dataset generated from experimental results obtained from literature and
    simulations performed with FLOW-3D. The corresponding melt pool dimensions
    are used in the computation of defect criterions used as classifications
    labels for a specific process parameter combination. A prompt which
    incorporates the process parameters is provided as input to the model which
    in turn predicts the potential defect classification.
  }
  \label{fig:main}
\end{figure}

\section{Methodology}
\subsection{Task}

Given an input text sequence outlining a L-PBF build process with descriptions
of process parameters such as material, power, and velocity; AdditiveLLM aims to
predict the most appropriate combination of defect classifications that is
likely to occur. The defect classifications of \textit{Keyholing}, \textit{Lack
of Fusion}, \textit{Balling}, and \textit{None} are one-hot encoded and provide
a set of labels for the multi-label classification task. The \textit{Baseline}
(Figure \ref{fig:baseline_template}) and \textit{Prompt} (Figure
\ref{fig:prompt_template}) datasets provide structure to the input in the form
of order dependent process parameters delineated with \texttt{[SEP]} separation
tokens or queries regarding process parameters in the format of natural language
respectively. The \textit{Prompt} dataset aims to capture the behavior of the
unstructured input expected from a user while the \textit{Baseline} dataset aims
to definitively capture each process parameter. These distinct approaches allow
for a more flexible use case for the models as the user is able to specifically
tune inputs specific to their use case.

\subsection{Dataset}

The dataset is compiled from literature and simulation values that provide
either melt pool dimensional measurements or corresponding defect
classifications from a set of process parameters. These process parameters
include \texttt{material}, \texttt{power}, \texttt{velocity},
\texttt{beam\_diameter}, \texttt{hatch\_spacing}, and \texttt{layer\_height} and
their corresponding units. The majority of the data only consists melt pool
dimensional measurements and these data points are processed into defect
classifications using estimations for \textit{Keyholing} (Eq. \ref{eq:keyhole}),
\textit{Lack of Fusion}(Eq. \ref{eq:lof}), and \textit{Balling} (Eq.
\ref{eq:balling}).

\begin{figure}[ht]
  \centering
  \includegraphics[width=\textwidth]{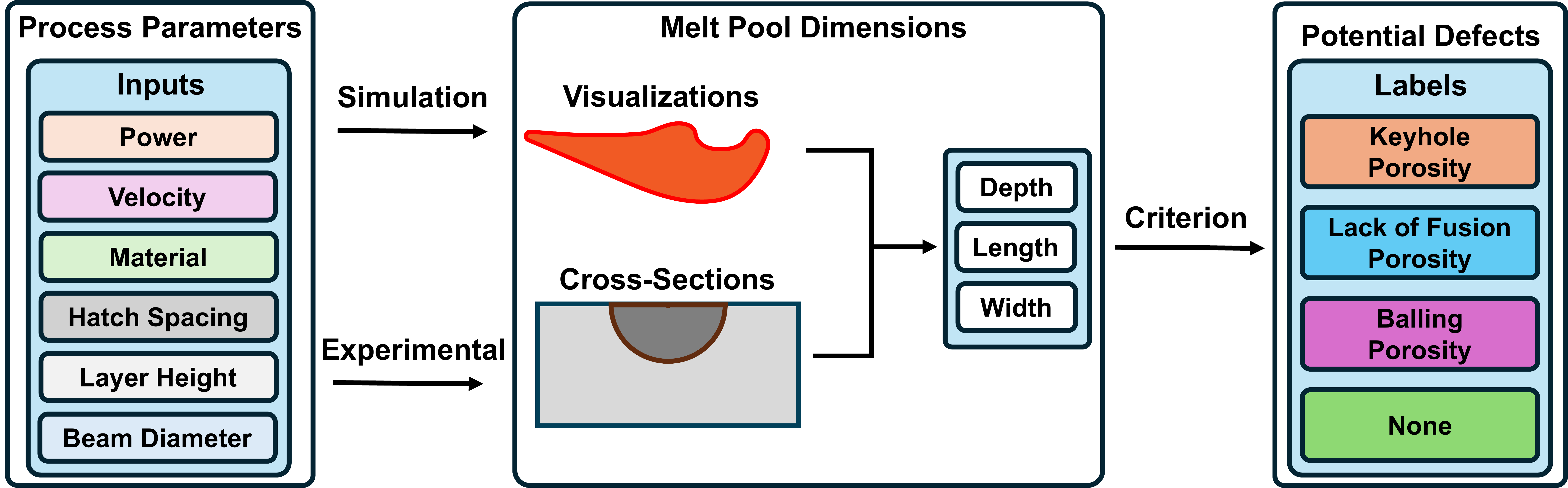}
  \caption{
    Melt pool dimensions dependent on the corresponding process parameters are
    obtained through either experimental cross-sectional measurements or
    computational fluid dynamic simulations. The obtained depth, length, and
    width, dimensions, pass through a set of criterions that determine the
    potential defects that would arise from the combination of process
    parameters.
  }
  \label{fig:dataset}
\end{figure}

A number of sources were used for our dataset which include literature and
simulation data points curated by \textit{Akbari et al.} in
\textit{MeltpoolNet}\cite{akbari_meltpoolnet_2022} and our own simulations
obtained using FLOW-3D\cite{noauthor_flow-3d_nodate}. (Figure \ref{fig:dataset})
This amounts to a total of 2779 collected data points spanning over 20 different
materials, primarily consisting of Ti-6Al-4V and Stainless Steel 316L. With
these sources, the dataset is comprised of over 20 different materials varying
with powers and velocity values ranging from 25 W to 5000 W and 1 mm/s to 2000
mm/s respectively.

The models were fine-tuned using this data using one hot encoded labels to
denote classifications of \textit{Keyhole}, \textit{Lack of Fusion},
\textit{Balling}, and \textit{None}. The label is formatted to be compatible
with a multi-label classification task since these defect regimes are not
mutually exclusive to one another. An example of two valid labels would be in
the case of a melt pool in keyhole mode creating pores while still producing
lack of fusion defects due to excessive layer height. The text input for the
dataset is formatted in two separate approaches, the first being
\textit{Baseline} (Figure \ref{fig:baseline_template}) structure relying on the
bare minimum amount of data and the second as the \textit{Prompt} (Figure
\ref{fig:prompt_template}) arrangement resembling inputs more akin to natural
language.

\subsubsection{Data Generation and Split}
The dataset consists of simulation and experimental data derived from sources
such as \textit{MeltpoolNet}\cite{akbari_meltpoolnet_2022}
(\texttt{meltpoolclassification.csv}, \texttt{meltpoolgeometry.csv}) and a
number of our own FLOW-3D\cite{noauthor_flow-3d_nodate} simulations
(\texttt{dimensions\_ti64.pkl} and \texttt{dimensions\_ss316l\_10\_micron.csv}).
The \texttt{meltpoolclassification.csv} provides the direct defect
classification of a combination of process parameters, however in cases where
only the melt pool dimensions are provided, the defect classifications are
obtained using their respective criterions. This data was further augmented by
analyzing the \textit{Lack of Fusion} defect criterion on 20 different layer
heights and hatch spacings ranging from 0 $\mu m$ to 950 $\mu m$ with the
assumption that the melt pool's width and depth remain the same. Each of these
individual files, all of the entries are shuffled and split to 75\%, 15\%, and
10\% for the train, test, and validation datasets respectively.

\subsubsection{\textit{Baseline} Dataset}

The \textit{Baseline} configuration of our dataset seeks to provide the minimum
input data necessary to fine-tune each model. This is done by utilizing the
\texttt{[SEP]} token to indicate the distinction between ordered process
parameters of \texttt{material}, \texttt{power}, \texttt{velocity},
\texttt{beam\_diameter}, \texttt{hatch\_spacing}, and \texttt{layer\_height}
(Figure \ref{fig:baseline_template}). The streamlined \texttt{text} input of
just the process parameters aims to provide the model only the essential
parameters to generate a prediction. This amounted to a total of 724,764 distinct
input-label pair combinations with the 75\%, 15\%, 10\% train, test, validation
split including 543,573, 108,715, and 72,476 respectively.

\subsubsection{\textit{Prompt} Dataset}

The \textit{Prompt} dataset consists of the same process parameter inputs but
structured in a way that more closely resembles natural language. This was
achieved through the use of ChatGPT to generate 100 unique prompt templates
querying for the potential defect associated with a combination of process
parameters. The template provides the same fields as the \textit{Baseline}
configuration to generate a set of text inputs for each set of process parameter
combinations within the data (Figure \ref{fig:prompt_template}). With the 75\%,
15\%, and 10\% respectively set to the train, test, and validation split, the
100 unique prompt templates are also split accordingly such that there are
either 75, 15, or 10 unique prompts generated from each set of process
parameters depending on the split of the dataset. This increases the input-label
pair of the dataset by 100x to over 70 million distinct pairs over all train,
test, and validation splits.

\subsection{Large Language Models}

\subsubsection{DistilBERT}
DistilBERT is a distilled implementation of the BERT model for the purposes of
fine tuning developed by researchers at HuggingFace\cite{sanh_distilbert_2020}.
DistilBERT achieves a 40\% reduction in model size (66 million parameters)
compared to the original BERT (110 million parameters) model while retaining
97\% of the language understanding capability and gaining a 60\% increase in
training speed\cite{sanh_distilbert_2020}. This model is a suitable candidate
for obtaining preliminary insights and evaluating the feasibility of
implementing a BERT-based inference approach for process parameter dependent
defects. The training process for the classification head a batch-size of 512
and a learning rate of 2E-5.

\subsubsection{SciBERT}
SciBERT is a specialized implementation of the 110 million parameter BERT
model\cite{devlin_bert_2019} pretrained on large-scale scientific data including
numerous publications from computer science to biomedical research
\cite{beltagy_scibert_2019}. With a dataset of over 1.14 million papers, the
authors are able to pretrain the BERT implementation on a corpus of text with
the use of domain specific vocabulary allowing better representation of
scientific terms and symbols through their SciVocab vocabulary library
\cite{beltagy_scibert_2019}. When evaluated on Natural Language Processing (NLP)
tasks regarding Named Entity Recognition (NER), PICO Extraction (PICO), and
Relation Classification (REL); SciBERT surpasses State Of The Art (SOTA)
performance, establishing its proficiency in comprehending and executing
scientific tasks \cite{beltagy_scibert_2019}. For this reason, we believe that
defect classification would be a suitable task for a finetuned implementation of
SciBERT with our \textit{Baseline} and \textit{Prompt} datasets. This model's
classification head was fine-tuned with a batch size of 256 and a learning rate
of 2E-5.

\subsubsection{Llama 3}
Llama 3 is the latest iteration of LLMs developed by the AI team at
Meta\cite{grattafiori_llama_2024}. In comparison to the previous
implementations, Llama 3 is trained on a much larger dataset of 15 trillion
tokens, boasts a larger context window of up to 128K tokens, and consists of 405
billion parameters for its largest model \cite{grattafiori_llama_2024,
touvron_llama_2023, touvron_llama_2023-1}. For our implementations we utilize
Llama-3.2-1B, a 1 billion parameter implementation of the the Llama 3 model that
incorporates the logits of the 8 billion and 70 billion parameter Llama 3.1
models during pretraining in order to retain a majority of their performance
while reducing the hardware requirements for
finetuning\cite{noauthor_meta-llamallama-32-1b_2024}. This model was fine-tuned
with a batch size of 512 and a learning rate of 2E-5.

\subsubsection{T5}
Text to Text Transfer Transformer (T5) is an encoder-decoder large language
model developed by Google specializing in simple text to text tasks such as
translation, summarization, and sentiment analysis \cite{raffel_exploring_2023}.
Variations of this model extend up to 11 billion parameters and for our
experiment we utilize T5-Small\cite{noauthor_google-t5t5-small_2024} consisting
of 60 million parameters for its efficient training speed and performance on a
specialized task such as defect classification \cite{raffel_exploring_2023,
noauthor_google-t5t5-small_2024}. This model was fine-tuned with a batch size of
512 and a learning rate of 1E-4 following the guidelines outlined in the
documentation \cite{noauthor_t5_nodate}.

\subsection{Training}
Each of the LLM models were attached with a classification head and fine-tuned
using the \textit{Baseline} and the \textit{Prompt} dataset. In each case the
weights of the LLM model remain unfrozen and were trained along with the
classification head for up to 25 epochs for \textit{Baseline} fine-tuning and 2
epochs for \textit{Prompt} fine-tuning. Models fine-tuned on the \textit{Prompt}
dataset were done so for a shorter number of epochs due to the lengthy training
duration resulting from the dataset being 100x larger than that of the
\textit{Baseline}. Binary cross entropy was used as the loss function for this
multi-label classification task and the inputs were tokenized with each LLM's
respective tokenizer used in pretraining. The models were fine-tuned on various
machines with GPU resources of Nvidia RTX A6000 with 48 GiB of memory and Nvidia
RTX 2080Ti with 11 GiB of memory dependent on which where available. The
Llama-3.2-1B model could only be fine-tuned with Nvidia RTX A6000 due to its
large overhead of 1 billion parameters and also took the most around of time to
train.

% Due to the large number of
% input-label pairs within the \textit{Prompt} dataset, over 40 million within the
% training split, these models could only be trained for only 2 epochs.

\section{Results and Discussion}

\subsection{Model Accuracy}
\begin{table}[h!]
\centering
\caption{
    Accuracy metrics obtained from validation set inference on models fine-tuned
    with \textit{Baseline} dataset for 25 epochs and \textit{Prompt} dataset 2
    epochs.
}
\begin{threeparttable}
\begin{tabular}{|c|c|c|}
\hline
\textbf{LLM}    & \textbf{\textit{Baseline} Fine-Tuned} & \textbf{\textit{Prompt} Fine-Tuned}   \\ \hline
DistilBERT      & 88.41\%                               & 82.22\%                               \\ \hline
SciBERT         & 90.88\%                               & 81.40\%                               \\ \hline
T5              & 71.414\%                              & \textbf{88.13\%}                      \\ \hline
Llama           & \textbf{93.68\%}                      & 73.52\%\tnote{a}                      \\ \hline
\end{tabular}
\begin{tablenotes}
\item[a] Fine-tuned for 0.5 epochs instead of the full 2 epochs.
\end{tablenotes}
\end{threeparttable}
\label{tab:model_accuracy_values}
\end{table}

\subsubsection{\textit{Prompt} Fine-Tuned LLMs}
Each of the LLMs were fine-tuned for 2 epochs with the exception of Llama where
it was fine-tuned for only 0.5 epochs due to its lengthy training duration.
Within the \textit{Prompt} fine-tuned LLMs, the T5 model achieves the highest
accuracy with DistilBERT and SciBERT achieving similar but lower accuracy
results. Llama presents the lowest accuracy however this is with the caveat that
it was fine-tuned for only half the dataset due to time constraints. With
further training we would expect the achieved accuracy would be much higher,
similar to that seen within the \textit{Baseline} fine-tuned LLMs.

\subsubsection{\textit{Baseline} Fine-Tuned LLMs}
Initially each of the LLM models were fine-tuned for 5 epochs on the
\textit{Baseline} dataset and further extended to 25 epochs with 5 epoch
increments (Table \ref{tab:model_accuracy_values_per_epoch}) which resulted in
all models apart from T5 showing significant improvement in evaluation accuracy.
This was most prominently displayed in the Llama model where the extended
training cycles drastically improved the model's prediction accuracy, achieving
an accuracy of 93.68\%, the highest accuracy of all the trained models (Table
\ref{tab:model_accuracy_values}). Llama's higher accuracy is expected as its
parameter size is over 9x greater (1 billion) to the BERT models which it is
compared against (110 million) while taking also significantly more time to
train as well. Notably, all models displayed plateauing or decreasing validation
accuracy after 20 epochs indicating that further fine-tuning may not present
additional benefit.

\begin{table}[h!]
\centering
\begin{tabular}{|c|c|c|c|c|c|}
\hline
\textbf{LLM}    & \textbf{5 Epochs} & \textbf{10 Epochs}    & \textbf{15 Epochs}    & \textbf{20 Epochs}    & \textbf{25 Epochs}   \\ \hline
DistilBERT      & 84.22\%           & 87.75\%               & 87.75\%               & 88.29\%               & 88.41\%              \\ \hline
SciBERT         & 88.29\%           & 88.93\%               & 88.99\%               & 91.32\%               & 90.88\%              \\ \hline
T5              & 79.61\%           & 71.45\%               & 76.42\%               & 77.92\%               & 71.41\%              \\ \hline
Llama           & \textbf{89.18\%}  & \textbf{91.66\%}      & \textbf{92.94\%}      & \textbf{93.01\%}      & \textbf{93.68\%}     \\ \hline
\end{tabular}
\caption{
    Accuracy for models trained on \textit{Baseline} dataset over a series of
    25 epochs with 5 epoch increments.
}
\label{tab:model_accuracy_values_per_epoch}
\end{table}

Of the four models trained with the \textit{Baseline} dataset, the T5 model
displayed the least potential as its accuracy remained around 88\% throughout
all training epochs whereas all other models exhibited an improvement in
evaluation accuracy. One reason for this could be attributed to the use of the
T5-Small model which only consisted of 60 million parameters, the least of all
compared models \cite{raffel_exploring_2023}. Its lightweight architecture and
smaller dimensionality of 512 contribute to its lack of accuracy improvement
after lengthening the number of training epochs \cite{raffel_exploring_2023}.

Evaluating the performance of the \textit{Prompt} dataset on LLMs fine tuned on
the \textit{Baseline} dataset, all exhibit poor performance. The initial
expectation was that the large language models would be able to reason through
input prompts and inherently map parameter values to those seen within the
\textit{Baseline} dataset. However with the high amount of defect
misclassifications, the models seem to be highly dependent on the syntactical
structure of process parameter inputs (Figure \ref{fig:baseline_template}) of
the \textit{Baseline} dataset. Potentially adjusting the \textit{Baseline}
dataset to further include the names of each process parameter inside the
input could help inform the model to the context of each value.

\subsection{Principle Component Analysis}
With the application of Principle Component Analysis (PCA) the model's ability
to distinguish between process parameters and their corresponding defects is
more clearly defined. In PCA charts for LLM models fine-tuned with the
\textit{Baseline} datasets, a clear distinction can be defects for \textit{Balling}
and \textit{Keyhole} / \textit{Lack of Fusion}. Areas of \textit{Keyhole} and
\textit{Lack of Fusion} share a greater amount of overlap and one explanation for
this lies in the data augmentation procedure used to determine \textit{Lack of Fusion}
defects.

\begin{figure}[ht]
  \centering
  \includegraphics[width=\textwidth]{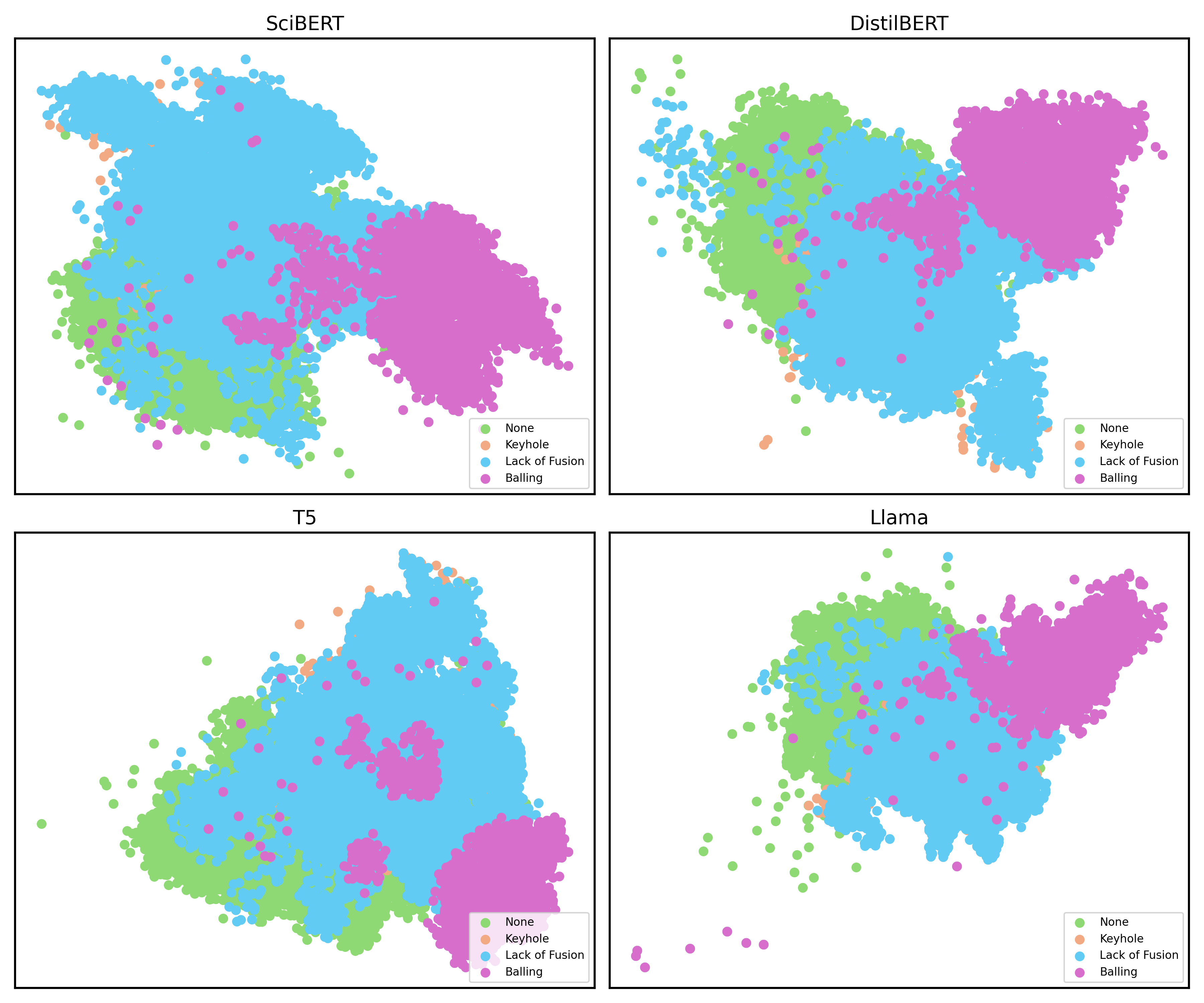}
  \caption{
    Principle Component Analysis (PCA) of LLMs fine-tuned (20 epochs) and
    evaluated with the \textit{Baseline} dataset. For all models, the labels of
    \textit{Keyhole} and \textit{Lack of Fusion} share a high degree of overlap
    to the extent that \textit{Lack of Fusion} covers much of the
    \textit{Keyhole} labels.
  }
  \label{fig:baseline_pca}
\end{figure}

\begin{figure}[ht]
  \centering
  \includegraphics[width=\textwidth]{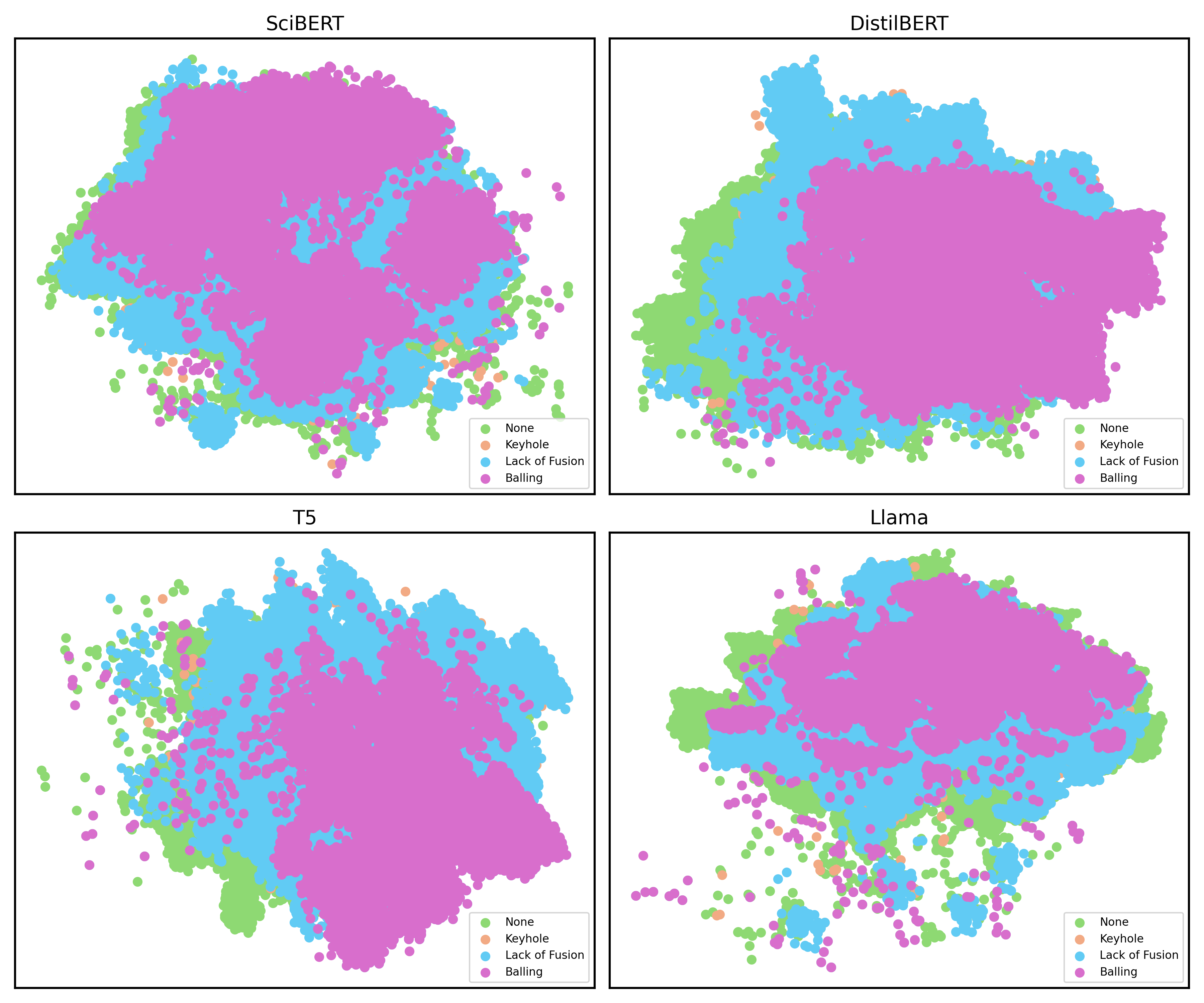}
  \caption{
    Principle Component analysis of LLM models fine-tuned on the
    \textit{Baseline} dataset and inferenced using the \textit{Prompt} dataset
    show high degrees of over between all labels indicating the models struggle
    to extract relevant information when formatted in natural language.
  }
  \label{fig:baseline_pca_prompt_input}
\end{figure}

The PCA chart regarding the \textit{Prompt} dataset (Figure
\ref{fig:baseline_pca_prompt_input}) exhibits a different trend where all of the
different defect regimes have high degrees of overlap on top of one another.
This further supports the case that the LLMs fine-tuned on the \textit{Baseline}
dataset struggle to interpret the process parameters outlined in the
\textit{Prompt} dataset, most likely due to the fact that process parameters are
not formatted in its expected syntax. As such, areas of \textit{Keyhole},
\textit{Lack of Fusion}, \textit{Balling}, and \textit{None} have high degrees
of overlap as the models fail to extract the significant features of each input.

\section{Conclusion and Future Work}

\subsection{Conclusion}
Analysis of fine-tuned LLM models show that the 1 billion parameter Llama 3
implementation achieves the highest accuracy of 93\% on the \textit{Baseline}
dataset and T5 displays the highest comparable accuracy within the
\textit{Prompt} fine-tuned LLMs. It is worth noting that the Llama model
presents a significantly longer training duration and parameter set than the
other LLMs. With this it can be seen that larger parameter sets and lengthier
training runs improve the evaluation accuracy of models fine tuned on the
\textit{Baseline} dataset. The findings here show that LLMs fine-tuned on a
process parameter defect dataset are able to reason and predict potential
defects existing in experimental and simulation data. The incorporation of
natural language allows for a more accessible inputs enabling users to determine
optimal build conditions with limited domain knowledge.

% In addition, inferences of the \textit{Prompt}
% dataset on models fine-tuned with the \textit{Baseline} dataset performed
% poorly, furthering the need to fine-tune these models with natural language
% formatted process parameters inputs in order to achieve accurate predictions.

\subsection{Future Work}
Future work in this domain would focus upon obtaining a more diverse dataset
with a wider range of materials, parameter sets, and additive processes. Longer
fine-tuning of LLMs with the \textit{Prompt} dataset could help expose the
models to diverse input set of natural language and produce predictions with
greater accuracy. In addition, parts of the existing models such as the
tokenizer could be replaced with one better equipped in understanding
numerical inputs.

\subsection{Data Availability}
The models and dataset are available within a HuggingFace collection at
% \url{https://huggingface.co/collections/ppak10/additivellm-6793c341ba5f8c22587a9d40}.
\href{https://additivellm.ppak.net}{additivellm.ppak.net}.

\begin{acknowledgement}

The author thanks Janghoon Ock for initial code reference and guidance on
general aspects for training and performing inference on Large Language Models
for domain specific tasks.

\end{acknowledgement}

\clearpage
\appendix

% Redefine \thesection to include "Appendix"
\renewcommand{\thesection}{Appendix \Alph{section}}

\section{Dataset Templates}
\begin{figure}[ht]
  \centering
  \includegraphics[width=\textwidth]{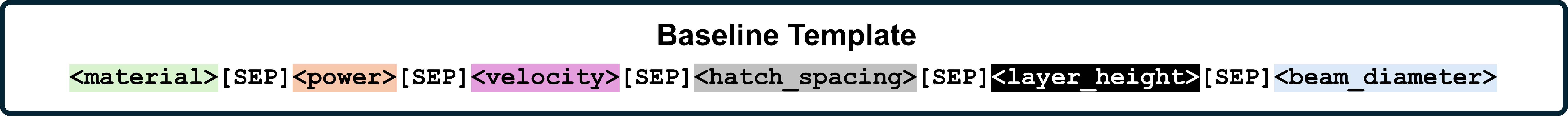}
  \caption{
    The highlighted attributes are replaced with the appropriate values and
    corresponding units, unknown values are replaced with an empty string.
  }
  \label{fig:baseline_template}
\end{figure}

\begin{figure}[ht]
  \centering
  \includegraphics[width=\textwidth]{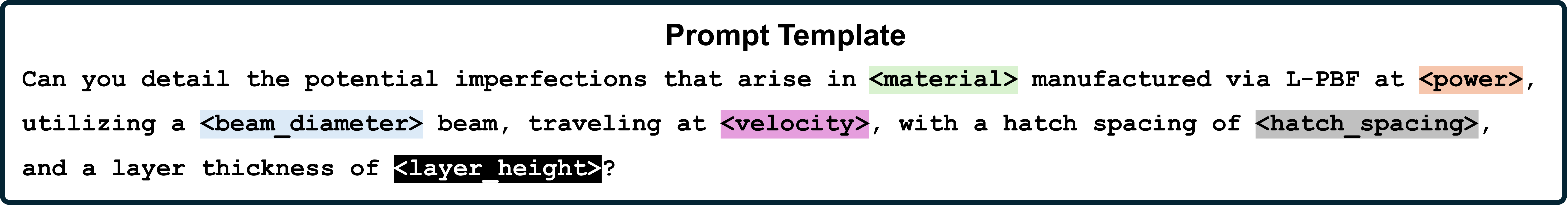}
  \caption{
    The text of each input label pair is formatted to a set of prompt templates,
    taking in consideration the multiple ways an input query can be structured.
  }
  \label{fig:prompt_template}
\end{figure}

% \bibliography{achemso-demo}
\bibliography{references}

\end{document}